\documentclass[10pt,twocolumn,letterpaper]{article}
\usepackage[paper]{cvpr}

\usepackage{subcaption} 
\usepackage{enumitem}
\usepackage[ruled,linesnumbered]{algorithm2e}
\DontPrintSemicolon

\definecolor{cvprblue}{rgb}{0.21,0.49,0.74}
\usepackage[pagebackref,breaklinks,colorlinks,allcolors=cvprblue]{hyperref}
\usepackage[capitalize]{cleveref}
\usepackage{multirow}
\usepackage{graphicx}
\usepackage{booktabs}
\usepackage{xcolor}
\usepackage{colortbl}

\definecolor{arcolor}{RGB}{0, 0, 0}
\definecolor{diffcolor}{RGB}{0, 0, 0}

\newcommand{\arbest}[1]{\textbf{\textcolor{arcolor}{#1}}}
\newcommand{\arsecond}[1]{\underline{\textcolor{arcolor}{#1}}}
\newcommand{\diffbest}[1]{\textbf{\textcolor{diffcolor}{#1}}}
\newcommand{\diffsecond}[1]{\underline{\textcolor{diffcolor}{#1}}}
\newcommand{\thename}{DiffusionVL}

\title{DiffusionVL: Translating Any Autoregressive Models into Diffusion \\Vision Language Models}

\author{
    Lunbin Zeng$^{1,*}$,
    Jingfeng Yao$^{1,*}$,
    Bencheng Liao$^{1}$,
    Hongyuan Tao$^{1}$,
    Wenyu Liu$^{1}$,
    Xinggang Wang$^{1,\dagger}$ \\
    $^{1}$Huazhong University of Science and Technology
}

\begin{document}
\twocolumn[{
  \renewcommand\twocolumn[1][]{#1}
  \vspace{-10pt}
  \maketitle
  \vspace{-10pt}
  \centering
  \includegraphics[width=\linewidth,keepaspectratio]{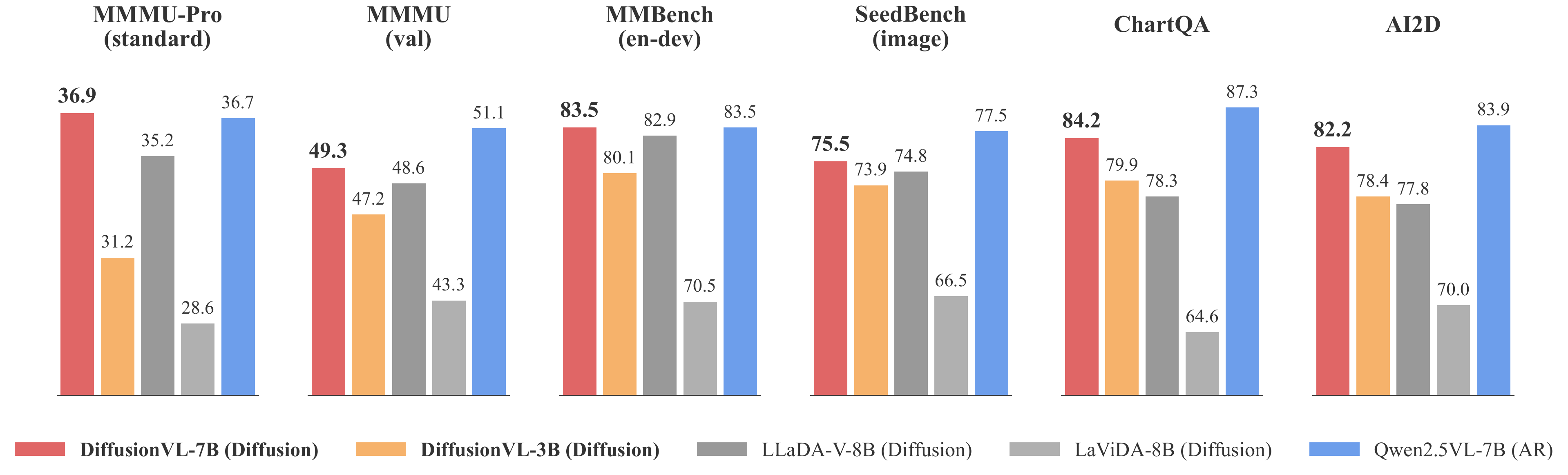}
  \captionof{figure}{\textbf{Performance Comparison.} Our DiffusionVL achieves state-of-the-art (SOTA) performance among diffusion vision language models including~\cite{li2025lavida, you2025llada} and competitive performance with Qwen2.5-VL~\cite{qwen2025qwen25technicalreport}.}
  \label{fig: benchmark}
     \vspace{20pt}
  }]
\maketitle

\begingroup
\makeatletter
\renewcommand\@makefnmark{}
\renewcommand\thefootnote{} 
\makeatother

\footnote{
    $^*$ Equal Contribution; 
    $^\dagger$ Corresponding author: Xinggang Wang 
    (\textcolor{blue}{\tt\small xgwang@hust.edu.cn}).
}

\addtocounter{footnote}{-1}
\endgroup

\begin{abstract}
    
Diffusion-based decoding has recently emerged as an appealing alternative to autoregressive (AR) generation, offering the potential to update multiple tokens in parallel and reduce latency. However, diffusion vision language models (dVLMs) still lag significantly behind mainstream autoregressive vision language models. This is due to the scarcity and weaker performance of base diffusion language models (dLLMs) compared with their autoregressive counterparts. This raises a natural question: Can we build high-performing dVLMs directly from existing powerful AR models, without relying on dLLMs?
We propose \textbf{\thename}, a family of dVLMs obtained by translating pretrained AR models into the diffusion paradigm via an efficient diffusion finetuning procedure that changes the training objective and decoding process while keeping the backbone architecture intact.
Through an efficient diffusion finetuning strategy, we successfully adapt AR pretrained models into the diffusion paradigm. 
This approach yields two key observations: (1) The paradigm shift from AR-based multimodal models to diffusion is remarkably effective. (2) Direct conversion of an AR language model to a dVLM is also feasible, achieving performance comparable to that of the same AR model finetuned with standard autoregressive visual instruction tuning. 
To enable practical open-ended generation, we further integrate block decoding, which supports arbitrary-length outputs and KV-cache reuse for faster inference. Our experiments demonstrate that despite training with \textbf{less than 5\% of the data} required by prior methods, \thename~achieves a comprehensive performance improvement, with \textbf{a 34.4\% gain} on the MMMU-Pro (vision) benchmark and \textbf{37.5\% gain} on the MME (Cog.) benchmark, alongside a \textbf{2× inference speedup}. The model and code are released at \url{https://github.com/hustvl/DiffusionVL}.

\end{abstract}
    
\section{Introduction}
\label{sec:intro}

Vision language models (VLMs)~\cite{bai2025qwen2,liu2023visual,chen2024internvl,tao2025infinitevlsynergizinglinearsparse} have achieved significant success in multimodal understanding. Most of them are autoregressive (AR) models with the next-token prediction (NTP) paradigm. A key limitation of the NTP paradigm is its inability to inherently support parallel inference, which limits its applicability in real-time scenarios. 

Recently, the diffusion paradigm~\cite{li2025lavida, you2025llada, yu2025dimple} has emerged as a promising alternative, offering more efficient parallel decoding potential than the NTP paradigm. 
However, existing diffusion VLMs (dVLMs) still lag behind advanced AR-VLMs on standard multimodal benchmarks (\cref{fig: benchmark}). 
A key bottleneck is that today's diffusion language models (dLLMs) are generally less capable and less mature than strong AR language models (AR-LMs), which limits the ceiling of dVLMs built on top of dLLMs.
For instance, LLaDA-8B lags behind Qwen2.5-7B by 42.0\% on the code task HumanEval \cite{chen2021evaluatinglargelanguagemodels}. Influenced by the conventional visual instruction tuning, existing attempts still assume that building dVLMs must rely on cross-modal training from a dLLM of the same paradigm. In fact, dVLMs and AR-VLMs are structurally identical; the only difference lies in their attention patterns and training/inference behaviors. 
Since the architecture does not dictate the paradigm, building dVLMs from dLLMs is not the only option. 
Given these challenges and findings, a compelling question emerges: \textit{Can we build high-performing dVLMs \emph{directly} from existing powerful AR models, without relying on dLLMs?}

\begin{figure}[t]
    \centering
    \includegraphics[width=\linewidth]{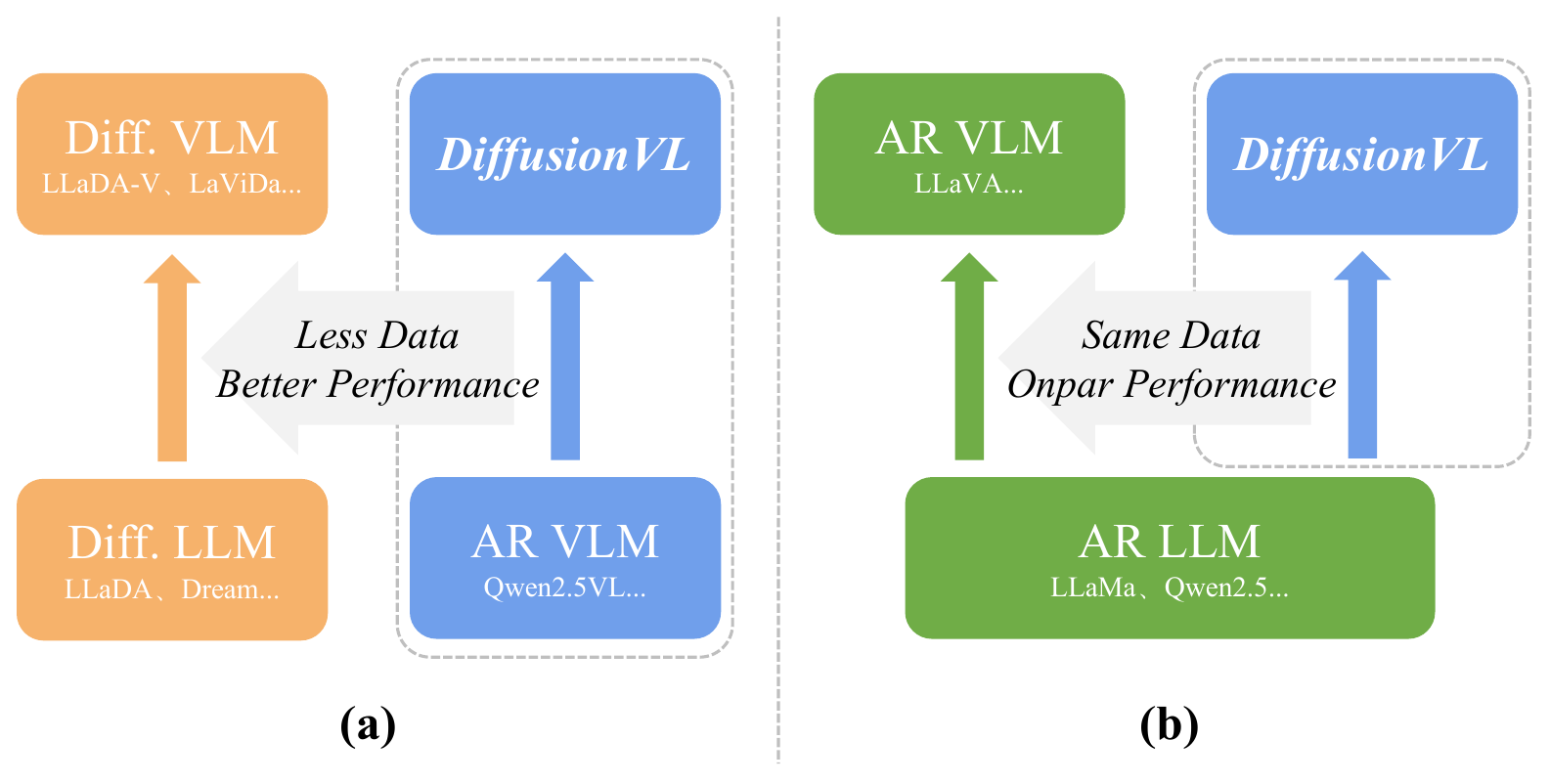}
    \caption{\textbf{Paradigm Shift and Modality Shift.} (a): translating AR-VLMs to dVLMs (paradigm shift only). (b): translating AR-LMs to dVLMs (modality shift and paradigm shift). We demonstrate that any autoregressive models with different modalities can be effectively translated to diffusion vision language models.}
    \label{fig: overview}
\end{figure}

To answer this question, we explore translating any pretrained autoregressive models into diffusion vision language models (see~\cref{fig: overview}). 
We propose \thename{}, whose core technical contribution is demonstrating that a simple diffusion finetuning approach can achieve this translation. 

Specifically, we convert the next-token prediction paradigm of the original AR model into a diffusion paradigm, which we refer to as "diffusion finetuning".
Its key advantage is that it enables the construction of \thename{} from any AR model without any architectural modifications. 
For AR-VLMs, since these models are already vision-language aligned, we directly apply full-parameter diffusion finetuning to convert AR-VLMs to dVLMs. 
For AR-LMs, we build the vision-language model in two stages. 
We first train only the connector during the pretraining stage to align the vision and text spaces, with the autoregressive paradigm ensuring training stability \cite{liu2023visual}.
We then conduct diffusion finetuning in the second stage to complete the paradigm conversion. Furthermore, we adopt the block decoding strategy consistent with \cite{arriola2025block}, which not only supports response generation of arbitrary lengths but also effectively reuses KV-cache for enhanced efficiency. 
Building on these technical designs, we introduce the \thename{} family, which can be translated from AR models of any scale and modality, while maintaining efficient inference speeds.

Our large amounts of experimental results validate the performance and efficiency of \thename{}. By seamlessly translating AR-VLMs into \thename{}, our method achieves state-of-the-art (SOTA) performance among current dVLMs. Remarkably, this is accomplished using less than 5\% of the training data required by prior approaches, thereby significantly narrowing the performance gap with advanced, data-intensive AR-VLMs. Specifically, \thename{} achieves a comprehensive performance improvement, with a 34.4\% gain on the MMMU-Pro (vision) benchmark and 37.5\% gain on the MME (Cog.) benchmark. When translating AR-LMs to DiffusionVL, we have drawn the following conclusions through detailed controlled experiments: under the same training data and configuration, the \thename{} finetuned from AR-LM consistently outperforms its counterpart finetuned from dLLM on downstream multimodal benchmarks. Furthermore, even when compared to AR-VLMs with autoregressive finetuning of the same paradigm, our \thename{} can still achieve comparable performance on downstream benchmarks. For inference, equipped with the block decoding strategy, \thename{} achieves a $2.0 \times$ speedup over previous dVLMs, demonstrating great potential of our method.

In summary, our contributions are threefold. 

\begin{itemize}
    \item We validate the effectiveness and feasibility of translating any pretrained autoregressive models into diffusion vision language models, providing an efficient and low-cost approach for developing high-performance diffusion vision language models. 
    \item We incorporate a block decoding strategy that enables arbitrary-length generation and efficient KV-cache reuse, addressing two key limitations of existing diffusion vision language models.
    \item Extensive experimental results demonstrate that our method yields a state-of-the-art \thename{}, which not only narrows the performance gap with advanced autoregressive vision language models but also achieves a $2.0\times$ speedup compared to existing diffusion vision language models.
\end{itemize}

\section{Related Work}

\label{sec:related}
\subsection{Masked Diffusion Models}

Diffusion models have achieved great success in vision generation and other computer vision tasks~\cite{peebles2023scalable,esser2024scaling,yao2025reconstruction,yao2024fasterdit,zou2025turbo,liao2025diffusiondrive,li2025recogdrive, Zhu_2025_CVPR}. At the same time, recent work has begun to explore the potential of diffusion in text generation. Prior work \cite{gong2024scaling,austin2021structured,lou2023discrete} on masked discrete diffusion models (MDMs) has already validated the effectiveness of this paradigm in small-scale text pretraining scenarios. These early studies demonstrated that MDMs can achieve perplexity levels comparable to those of AR-LMs while enabling parallel inference. Building on this basis, recent advanced models such as LLaDA \cite{nie2025large} and Dream \cite{ye2025dream} have further validated the effectiveness and scalability of MDMs in modern large language model (LLM) settings. 

In the visual domain, diffusion vision language models (dVLMs) are typically built from a pre-trained diffusion language model and converted through visual instruction finetuning. Dimple \cite{yu2025dimple} designs a novel training paradigm that combines an autoregressive phase with a subsequent diffusion phase. LLaDA-V \cite{you2025llada} directly explores the effectiveness of large-scale visual finetuning under the diffusion paradigm based on LLaDA. LaViDa \cite{li2025lavida} further introduces complementary masking and prefix cache to improve training and inference efficiency. These models follow the LLaVA \cite{liu2023visual} paradigm and have explored the potential of the mask diffusion paradigm in the visual and multimodal domains. However, they are limited by their inability to perform variable-length generation and to reuse the KV-cache efficiently. A notable performance gap also remains compared with advanced AR-VLMs. Beyond multimodal understanding, recent work extends the masked diffusion paradigm to unified frameworks: MMaDA \cite{yang2025mmada} and LaViDa-O \cite{li2025lavidaoelasticlargemasked} demonstrate that the masked diffusion paradigm, when combined with finetuning and reinforcement learning, can support both image and text generation and editing within a single model.

\subsection{Interpolation between AR and Diffusion}

Due to limitations of the MDMs in KV-cache reuse and arbitrary-length generation, several works have explored how to combine the autoregressive (AR) and masked diffusion paradigms for text generation. SSD-LM \cite{han2022ssd} introduced a block formulation of Gaussian text diffusion. AR-Diffusion \cite{wu2023ar} further extended SSD-LM by incorporating a left-to-right noise schedule. BD3-LM \cite{arriola2025block} interpolates between discrete masked diffusion and autoregressive models by using inner-block diffusion and inter-block autoregressive decoding, and has served as a foundation for follow-up research. However, these early block diffusion methods are limited to small-scale text pre-training; their scalability to large language models and extension to the vision-language domain remained unproven.

Recently, some works have attempted to use autoregressive models as initialization and convert them into block diffusion paradigm. To scale this block diffusion paradigm to large text models, SDAR \cite{JetAstra2025}, Fast-dLLM-V2 \cite{wu2025fastv2} verified that finetuning from AR models can build efficient block diffusion language models. SDLM \cite{liu2025sequential} further extended the block diffusion paradigm to next-sequence prediction. More recently, LLaDA~2.0 \cite{bie2025llada20scalingdiffusionlanguage} scales block diffusion to 100B parameters and optimizes inference speed, achieving performance competitive with advanced open-source AR-LMs while delivering significantly faster inference. However, such AR-diffusion interpolation remains underexplored in vision language models; our work addresses this gap by translating any autoregressive models into dVLMs through a simple diffusion finetuning. 

\section{Method}
\label{sec:method}
\subsection{Preliminaries}
\label{sec: preliminary}

In the autoregressive paradigm, text generation is modeled as next-token prediction. Given an input sequence $\{x^1,\dots,x^L\}$, the training objective is to minimize the cross-entropy loss. 
\begin{equation}
\label{eq:ar_loss}
\mathcal{L}_{\text{AR}}(x; \theta) = -\mathbb{E}_x \left[ \sum_{i=1}^L \log P_\theta(x^i \mid x^{<i}) \right],
\end{equation}
where the model $P_\theta(\cdot |x^{<i})$ aims to maximize the conditional probability of the current position by using the preceding context $x^{<i}=x^1,\dots,x^{i-1}$.

By contrast, masked diffusion models can be subdivided into two paradigms: full diffusion and block diffusion. The full diffusion paradigm adds and removes noise simultaneously throughout the entire sequence. For a time $t \in(0,1)$, the input sequence is masked with a probability of $t$, yielding a noisy sequence $x_t$. The model $P_\theta(\cdot |x_t)$ is trained to minimize the expected masked position prediction loss. 
\begin{equation}
\label{eq:d_loss}
\mathcal{L}_{\text{DM}}(x; \theta) = -\mathbb{E}_{t,x_0,x_t} \left[ \frac{1}{t} \sum_{i \in \mathcal{M}_t} \log P_\theta(x_0^i \mid x_t) \right],
\end{equation}
where $t \sim \mathcal{U}(0, 1)$ and $\mathcal{M}_t$ denotes the set of masked positions.

The block diffusion paradigm divides the entire sequence into several blocks of equal size and performs block-wise noise addition and denoising within each block.
\begin{equation}
\label{eq:bd_loss}
\mathcal{L}_{\text{BDM}}(x; \theta) = -\mathbb{E}_{t,x_0,x_t} \left[ \sum_{i \in \mathcal{M}_t} \alpha \,\log P_\theta(x_0^i \mid \mathcal{C}_i) \right],
\end{equation}
where $\mathcal{C}_i=(x_{<i},x_{D(i)})$, $x_{<i}$ are clean contexts from earlier blocks, $x_{D(i)}$ denotes all contexts in the block containing position $i$, and $\alpha=\alpha_t'/(1-\alpha_t)$ with $\alpha_t'$ the instantaneous rate of change of $\alpha_t$ in continuous time.

\subsection{Modalities and Paradigms}
\label{sec: modalities and paradigms}

\begin{figure*}[t]
    \centering
    \includegraphics[width=0.9\linewidth]{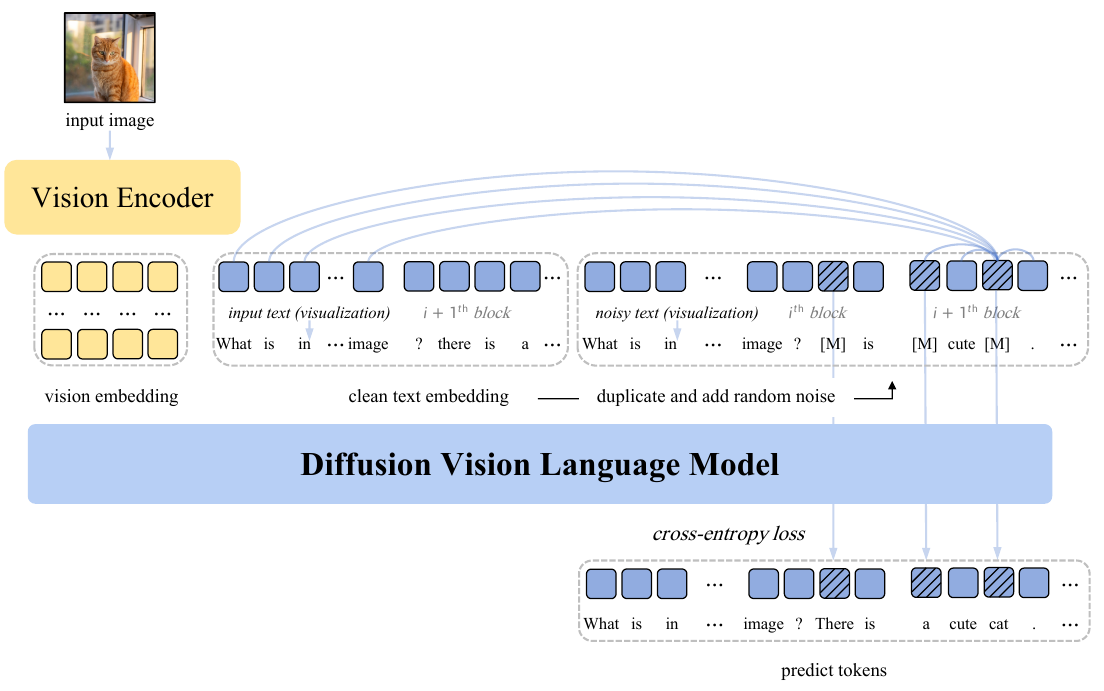}
    \caption{\textbf{The diffusion finetuning framework of DiffusionVL.} The noisy sequence is concatenated with the original clean sequence along the sequence dimension. Each noisy block attends to all preceding clean blocks (inter-block causal) and all positions within its own block (intra-block bidirectional). The cross-entropy loss is computed only at masked positions.}
    \label{fig: train}
\end{figure*}

Architectures and paradigms are largely decoupled: A single Transformer architecture \cite{vaswani2017attention} can yield diverse models by applying different paradigms. Based on this observation, our model retains the same network architecture as existing autoregressive models and is initialized from them; only the training/inference behavior and the attention pattern differ. Accordingly, we design different diffusion finetuning pipelines for AR-VLMs and AR-LMs.

First, we describe the process of finetuning an AR-VLM into a dVLM as a paradigm shift. We demonstrate that this direct paradigm shift is an efficient way to build dVLMs, which aligns with the findings of \cite{JetAstra2025,wu2025fastv2} in the text domain. Since our model has already been vision-language aligned, we directly finetune the entire AR-VLM into a dVLM in an end-to-end manner using Eq.~\eqref{eq:bd_loss}.

Furthermore, we extend the approach to build a dVLM based on an AR-LM, and describe it as a process of modality shift and paradigm shift. We adopt the traditional two-stage training approach similar to LLaVA \cite{liu2023visual}. In the pretraining stage, we only train the connector to align the vision embedding space and text embedding space. Since this connector is randomly initialized to align vision and text embeddings, we use the standard autoregressive objective Eq.~\eqref{eq:ar_loss} to stabilize the modality shift process. In the finetuning stage, all components are jointly trained end-to-end using the diffusion finetuning strategy Eq.~\eqref{eq:bd_loss} to achieve modality shift and paradigm shift at the same time.

\subsection{Diffusion Finetuning}
\label{sec: diffusion finetuning}

In this section, we elaborate on the training framework of our diffusion finetuning, as illustrated in~\cref{fig: train}. 

The image is processed through the vision encoder and projector, and the resulting vision embeddings are concatenated with text embeddings. Each sequence is padded with \texttt{<EOS>} tokens to a length divisible by the block size $D$ and split into $B=L/D$ non-overlapping blocks $\{\mathbf{x}_{(1)}, \ldots, \mathbf{x}_{(B)}\}$. Unlike the sequence-level noise in prior dVLMs \cite{you2025llada,li2025lavida,yu2025dimple}, we adopt a block-wise noise schedule: For each block $b$ containing response or padding tokens, a noise level $t_b \sim \mathcal{U}(0,1)$ is sampled and each token $x^i$ within it is independently masked as
\begin{equation}
\label{eq:block_noise}
\tilde{x}^i=
\begin{cases}
\texttt{[MASK]}, & p=t_b, \\
x^i, & p=1-t_b,
\end{cases}
\end{equation}
where $p$ is the noise probability, while blocks containing image or prompt tokens remain unmasked ($t_b = 0$). The noised sequence $\tilde{\mathbf{x}}$ and clean sequence $\mathbf{x}$ are concatenated along the sequence dimension, with an attention mask $\mathbf{M}$ enforcing intra-block bidirectional and inter-block causal attention. For position $i$ in noised block $b$, $\mathbf{M}_{ij}$ indicates whether $i$ may attend to $j$ ($1$) or not ($0$):
\begin{equation}
\label{eq:attn_mask}
    \mathbf{M}_{ij} = 
    \begin{cases}
        1, & \text{if } j \text{ is in the same noised block as } i, \\
        1, & \text{if } j \text{ is in a clean block $b'$ with $b' < b$,} \\
        0, & \text{otherwise}.
    \end{cases}
\end{equation}
The model is trained to minimize the cross-entropy loss at masked positions using Eq.~\eqref{eq:bd_loss}.

\subsection{Diffusion Inference}
\label{sec: diffusion inference}

\begin{algorithm}[t]
    \small
    \setlength{\algomargin}{1.5em}
    \caption{Diffusion Inference of \thename.}\label{alg:inference}
    \KwIn{image $\mathbf{I}$, prompt $\mathbf{T}$, max length $L$, block size $D$, steps $S$, threshold $\tau$}
    \KwOut{generated token sequence $\mathbf{Y}$}
    $\mathbf{C}_0\gets \texttt{cat}\bigl(\texttt{proj}(\texttt{ve}(\mathbf{I})),\texttt{embed}(\mathbf{T})\bigr);\;\mathbf{Y}\gets\varnothing$\;
    \For{$m \gets 1$ \KwTo $\lfloor L/D \rfloor$}{
        $\mathbf{x}_m^{(0)} \gets [\texttt{MASK}]^D$\;
        \For{$s \gets 1$ \KwTo $S$}{
            $\hat{x}_{m,i}^{(s)} \sim P_\theta(\cdot \mid \mathbf{x}_m^{(s-1)},\, \mathbf{C}_{m-1})$,\; $c_{m,i}^{(s)} \gets P_\theta(\hat{x}_{m,i}^{(s)} \mid \mathbf{x}_m^{(s-1)},\, \mathbf{C}_{m-1})$\; \textbf{for all} $i \in \mathcal{M}^{(s-1)}$\;
            $k \gets \min(\lceil D/S \rceil, |\mathcal{M}^{(s-1)}|)$;\; $\mathcal{U}^{(s)} \gets \operatorname{top\text{-}k}(\{c_{m,i}^{(s)}\}, k)$\;
            \lIf{\textnormal{dynamic}}{$\mathcal{U}^{(s)} \gets \mathcal{U}^{(s)} \cup \{i \in \mathcal{M}^{(s-1)} \mid c_{m,i}^{(s)} > \tau\}$}
            $x_{m,i}^{(s)} \gets \hat{x}_{m,i}^{(s)}$ if $i \in \mathcal{U}^{(s)}$, else $\texttt{[MASK]}$\;
        }
        $\mathbf{C}_m \gets \mathbf{C}_{m-1} \cup \mathbf{x}_m^{(S)};\; \mathbf{Y} \gets \mathbf{Y} \cup \mathbf{x}_m^{(S)}$\;
        \lIf{$\langle\textup{EOS}\rangle \in \mathbf{x}_m^{(S)}$}{\textbf{break}}
    }
    \Return{$\mathbf{Y}$}
\end{algorithm}

During inference, our model combines inter-block autoregressive generation with intra-block parallel diffusion decoding, naturally supporting KV-cache reuse and arbitrary-length generation. The full procedure is summarized in~\cref{alg:inference}. 

Given an input image and a text prompt, we first encode them using the vision encoder, projector, and text embedding layer, and concatenate the resulting embeddings to initialize the context cache $\mathbf{C}_0$. For each subsequent block $m$, we initialize all $D$ positions as mask tokens and perform up to $S$ iterative denoising steps. 

At each step $s$, let $\mathcal{M}^{(s-1)}$ denote the set of indices still masked in $\mathbf{x}_m^{(s-1)}$. For each $i \in \mathcal{M}^{(s-1)}$, we sample the prediction conditioned on the current block state and preceding context $\mathbf{C}_{m-1}$ via KV-cache reuse (following the hybrid attention pattern in Eq.~\eqref{eq:attn_mask}), and record the confidence $c_{m, i}^{(s)}$. A subset $\mathcal{U}^{(s)}$ is then chosen and unmasked based on $c_{m,i}^{(s)}$. 

We adopt two remasking strategies following \cite{JetAstra2025} to determine $\mathcal{U}^{(s)}$: the static low-confidence remasking strategy unmasks the $k$ highest-confidence positions per step; the dynamic low-confidence remasking strategy further augments this set by additionally unmasking all positions whose confidence exceeds a threshold $\tau$, enabling faster decoding on simpler content. Once a block is fully denoised, it is appended to the context cache, and generation proceeds to block $m+1$ until an $\langle\text{EOS}\rangle$ token is produced.

\section{Experiment}

\begin{table*}[t]
\centering
\renewcommand{\arraystretch}{1.2}
\resizebox{\textwidth}{!}{
\begin{tabular}{l|ccc|ccccccc}
\toprule
\textbf{Model} & \textbf{Size} & \textbf{Type} & \textbf{Samples} & 
\textbf{\begin{tabular}{@{}c@{}}MMBench\\ {[en-dev]}\end{tabular}} & 
\textbf{\begin{tabular}{@{}c@{}}MMMU\\ {[val]}\end{tabular}} & 
\textbf{\begin{tabular}{@{}c@{}}MMMU-Pro\\ {[std.]}\end{tabular}} & 
\textbf{\begin{tabular}{@{}c@{}}MMMU-Pro\\ {[vision]}\end{tabular}} & 
\textbf{\begin{tabular}{@{}c@{}}MMStar\\ {[test]}\end{tabular}} & 
\textbf{\begin{tabular}{@{}c@{}}MME\\ {[cog.]}\end{tabular}} & 
\textbf{\begin{tabular}{@{}c@{}}MME\\ {[perp.]}\end{tabular}} \\
\midrule
\multicolumn{11}{c}{\textbf{\textit{AutoRegressive Vision Language Models}}} \\
\midrule
\textbf{LLaVA} & 7B & AR & - & 38.7 & - & - & - & - & - & 809 \\
\textbf{LLaVA-1.5} & 7B & AR & - & 64.3 & - & - & - & - & - & 1510 \\
\textbf{Cambrian-1} & 8B & AR & - & 75.9 & 42.7 & - & - & - & - & 1547 \\
\textbf{LLaVA-OV} & 7B & AR & 7.8M & \arsecond{80.8} & \arsecond{48.8} & - & - & \arsecond{61.7} & 418 & \arsecond{1580} \\
\multirow{2}{*}{\textbf{Qwen2.5VL}} & 3B & AR & \textgreater 9M & 79.1 & 46.8 & \arsecond{31.2} & \arsecond{22.1} & 55.9 & \arsecond{620} & 1533 \\
 & 7B & AR & \textgreater 9M & \arbest{83.5} & \arbest{51.1} & \arbest{36.7} & \arbest{33.4} & \arbest{63.9} & \arbest{646} & \arbest{1680} \\
\midrule
\multicolumn{11}{c}{\textbf{\textit{Diffusion Vision Language Models}}} \\
\midrule
\textbf{LaViDa-L} & 8B & Diff. & 1.6M & 70.5 & 43.3 & 28.6 & - & - & 341 & 1365 \\
\textbf{Dimple} & 7B & Diff. & 1.3M & - & 45.2 & - & - & - & 432 & 1514 \\
\textbf{LLaDA-V} & 8B & Diff. & 16.5M & \diffsecond{82.9} & 48.6 & \diffsecond{35.2} & 18.6 & \diffsecond{60.1} & 491 & 1507 \\
\multirow{2}{*}{\textbf{\thename{}}} & 3B & Diff. & 738K & 80.1 & 47.2 & 31.2 & \diffsecond{20.2} & 55.9 & \diffsecond{594} & \diffbest{1539} \\
 & 7B & Diff. & 738K & \diffbest{83.5} & \diffbest{49.3} & \diffbest{36.9} & \diffbest{25.0} & \diffbest{63.2} & \diffbest{675} & \diffsecond{1519} \\
\bottomrule
\end{tabular}
}
\caption{\textbf{Benchmark Performance Comparison (Part 1)}. The top-2 results are highlighted separately for AR and Diffusion models. Best results are in \textbf{bold}, second-best are \underline{underlined}.}
\label{tab:table1_part1_colored}
\end{table*}

\begin{table*}[t]
\centering
\renewcommand{\arraystretch}{1.2}
\resizebox{\textwidth}{!}{
\begin{tabular}{l|ccc|cccccc}
\toprule
\textbf{Model} & \textbf{Size} & \textbf{Type} & \textbf{Samples} & 
\textbf{\begin{tabular}{@{}c@{}}SeedBench\\ {[img]}\end{tabular}} & 
\textbf{\begin{tabular}{@{}c@{}}SeedBench\\ {[vid]}\end{tabular}} & 
\textbf{AI2D} & 
\textbf{ChartQA} & 
\textbf{\begin{tabular}{@{}c@{}}Realworld\\ QA\end{tabular}} & 
\textbf{MuirBench} \\
\midrule
\multicolumn{10}{c}{\textbf{\textit{AutoRegressive Vision Language Models}}} \\
\midrule
\textbf{LLaVA} & 7B & AR & - & 37.0 & 23.8 & - & - & - & - \\
\textbf{LLaVA-1.5} & 7B & AR & - & 66.1 & 37.3 & - & - & - & - \\
\textbf{Cambrian-1} & 8B & AR & - & 74.7 & - & 73.0 & 73.3 & 64.2 & - \\
\textbf{LLaVA-OV} & 7B & AR & 7.8M & \arsecond{75.4} & \arsecond{56.9} & 81.4 & 80.0 & \arsecond{66.3} & 41.8 \\
\multirow{2}{*}{\textbf{Qwen2.5VL}} & 3B & AR & \textgreater 9M & 74.8 & 55.1 & \arsecond{81.6} & \arsecond{84.0} & 65.4 & \arsecond{47.7} \\
 & 7B & AR & \textgreater 9M & \arbest{77.5} & \arbest{61.3} & \arbest{83.9} & \arbest{87.3} & \arbest{68.5} & \arbest{59.6} \\
\midrule
\multicolumn{10}{c}{\textbf{\textit{Diffusion Vision Language Models}}} \\
\midrule
\textbf{LaViDa-L} & 8B & Diff. & 1.6M & 66.5 & - & 70.0 & 64.6 & - & - \\
\textbf{Dimple} & 7B & Diff. & 1.3M & - & - & 74.4 & 63.4 & - & - \\
\textbf{LLaDA-V} & 8B & Diff. & 16.5M & \diffsecond{74.8} & \diffsecond{53.7} & 77.8 & 78.3 & \diffsecond{63.2} & \diffbest{48.3} \\
\multirow{2}{*}{\textbf{\thename{}}} & 3B & Diff. & 738K & 73.9 & 52.2 & \diffsecond{78.4} & \diffsecond{79.9} & 61.6 & \diffsecond{47.2} \\
 & 7B & Diff. & 738K & \diffbest{75.5} & \diffbest{54.4} & \diffbest{82.2} & \diffbest{84.2} & \diffbest{68.0} & 44.8 \\
\bottomrule
\end{tabular}
}
\caption{\textbf{Benchmark Performance Comparison (Part 2)}. \thename-7B achieves top-tier performance among Diffusion models and closes the gap with top AR models. Best in \textbf{bold}, second-best \underline{underlined}.}
\label{tab:table1_part2_colored}
\end{table*}

\begin{table*}[t]
\centering
\renewcommand{\arraystretch}{1.2}
\setlength{\tabcolsep}{6pt}
\resizebox{\textwidth}{!}{
\begin{tabular}{l|c|c|ccccc}
\toprule
\textbf{Model} &
\textbf{Paradigm} &
\textbf{\begin{tabular}[c]{@{}c@{}}Base LLM\\(MMLU)\end{tabular}} &
\textbf{\begin{tabular}[c]{@{}c@{}}MMMU\\(val)\end{tabular}} &
\textbf{\begin{tabular}[c]{@{}c@{}}MMMU-Pro\\(std.)\end{tabular}} &
\textbf{ChartQA} &
\textbf{\begin{tabular}[c]{@{}c@{}}RealWorld\\QA\end{tabular}} &
\textbf{\begin{tabular}[c]{@{}c@{}}MME\\(cog.)\end{tabular}} \\
\midrule
\multicolumn{8}{c}{\textbf{\textit{Conversion from LLaDA-8B (dLLM base)}}} \\
\midrule
\textbf{LLaDA-V} & Full-Diff   & \multirow{2}{*}{65.9} & 42.4 & 26.0 & 20.2 & 60.1 & 342 \\
\textbf{LLaDA-V} & Block-Diff. & & 32.6 & 17.1 & 29.8 & 44.2 & 261 \\

\midrule
\multicolumn{8}{c}{\textbf{\textit{Conversion from Qwen2.5-7B (AR-LM base)}}} \\
\midrule
\textbf{LLaVA}       & AR          & \multirow{2}{*}{71.9} & \textbf{45.4} & \underline{28.1} & \underline{52.8} & \underline{60.4} & \underline{356} \\
\textbf{DiffusionVL} & Block-Diff. & & \underline{43.7} & \textbf{28.4} & \textbf{53.6} & \textbf{60.7} & \textbf{371} \\

\bottomrule
\end{tabular}
}
\caption{\textbf{A comparison of dVLM construction using different models and paradigms.} We compare dVLMs converted from AR-LMs versus those from dLLMs. The Base LLM column indicates MMLU scores of the base language models. The best results across all models are highlighted in \textbf{bold}, and the second-best are \underline{underlined}. The AR, Block-Diff, and Full-Diff paradigms represent the different paradigms we discussed in~\cref{sec: preliminary}.}
\label{tab: comparison of dVLM}
\end{table*}

\subsection{Implementation Details}
\label{sec: exp_set}

\noindent\textbf{Model architecture.} For building dVLMs from AR-VLMs, we use Qwen2.5-VL-3B-Instruct and Qwen2.5-VL-7B-Instruct~\cite{bai2025qwen2} as base models. For finetuning from AR-LMs and dLLMs, we select Qwen2.5-7B-Instruct~\cite{qwen2025qwen25technicalreport} and LLaDA-8B-Instruct~\cite{nie2025large}, respectively. The vision encoder is SigLip2-400M~\cite{tschannen2025siglip}. The projector is a randomly initialized two-layer MLP.

\noindent\textbf{Training data.} For building dVLMs from AR-VLMs, we use only the 738K instruction-following samples from LLaVA-Next~\cite{li2024llavanext-strong} for end-to-end finetuning. For building dVLMs and AR-VLMs from AR-LMs and for building dVLMs from dLLMs, we adopt a two-stage data setup: the 580K-sample pretraining dataset from LLaVA-Pretrain~\cite{liu2023visual} in the pretraining stage, and the 738K instruction-following samples from LLaVA-Next in the finetuning stage.

\noindent\textbf{Hyperparameters.} We use AdamW with a cosine decay scheduler. The pretraining stage uses a learning rate of $1\times10^{-3}$; the finetuning stage uses $1\times10^{-5}$ for the LLM and projector and $2\times10^{-6}$ for the vision encoder. The default training block size is~8. During inference, we adopt static low-confidence remasking with the number of denoising steps equal to the block size.

\noindent\textbf{Evaluation benchmarks.} We evaluate on a diverse set of vision-language benchmarks using the LMMS-Eval~\cite{zhang2024lmmsevalrealitycheckevaluation} library with its default prompts:
\begin{itemize}[leftmargin=*,nosep]
    \item \textit{General knowledge:} MMMU~\cite{yue2024mmmu}, MMMU-Pro~\cite{yue2024mmmupro}, MMStar~\cite{chen2024we}, MME~\cite{fu2025mmecomprehensiveevaluationbenchmark}, SeedBench~\cite{li2023seed}, MMBench~\cite{liu2024mmbench}, RealworldQA~\cite{xai2024grok15v}.
    \item \textit{Chart and document understanding:} AI2D~\cite{kembhavi2016diagram}, ChartQA~\cite{masry2022chartqa}.
    \item \textit{Mathematical reasoning:} MathVista~\cite{lu2023mathvista}, MathVerse~\cite{zhang2024mathverse}, MathVision~\cite{ahmad2024mathvision}.
    \item \textit{Detail image captioning:} DetailCaps~\cite{dong2024benchmarking}.
    \item \textit{Multi-image understanding:} MuirBench~\cite{wang2024muirbench}.
\end{itemize}

\subsection{Efficient dVLM Construction from AR-VLMs}
\label{sec: exp_p1}

\cref{tab:table1_part1_colored} and~\cref{tab:table1_part2_colored} present the downstream multimodal benchmark results of DiffusionVL-3B and DiffusionVL-7B, which are derived from diffusion finetuning of Qwen2.5VL-3B-Instruct and Qwen2.5VL-7B-Instruct. We compare our models' results with representative AR-VLMs and open-source dVLMs.

As shown in~\cref{tab:table1_part1_colored} and~\cref{tab:table1_part2_colored}, DiffusionVL-7B achieves comprehensive performance that surpasses existing open-source dVLMs LaViDa-L \cite{li2025lavida}, Dimple \cite{yu2025dimple}, and LLaDA-V \cite{you2025llada} on multimodal benchmarks. This is achieved despite finetuning with only 738K samples, i.e., less than 5\% of the data used for LLaDA-V, demonstrating strong vision-language capability and high training efficiency. Furthermore, leveraging the strong pretrained foundation of AR-VLMs, DiffusionVL-3B even outperforms the larger LaViDa-L-8B and Dimple-7B with less training data, further validating the effectiveness of our proposed method. Moreover, our DiffusionVL significantly narrows the gap between existing dVLMs and advanced AR-VLMs, demonstrating an efficient approach to building dVLMs.

\subsection{Feasible dVLM Conversion from AR-LMs}
\label{sec: exp_p2}

In this section, we conduct multiple groups of experiments on different base language models and various finetuning paradigms to investigate the feasibility of finetuning dVLMs from AR-LMs. We perform autoregressive finetuning, block diffusion finetuning, and full diffusion finetuning based on the AR-LM Qwen2.5-7B-Instruct and the dLLM LLaDA-8B-Instruct, respectively, and report the differences in their performance on the downstream multimodal benchmarks under the same training settings and data.

As shown in~\cref{tab: comparison of dVLM}, DiffusionVL significantly outperforms LLaDA-V (finetuned from LLaDA using block diffusion and full diffusion paradigms). This indicates that building a high-performance dVLM does not require a pre-existing dLLM; we can fully leverage existing powerful AR-LMs to develop such dVLMs. Notably, DiffusionVL exhibits negligible differences from LLaVA on downstream benchmarks. This further confirms that constructing dVLMs from AR-LMs is both feasible and effective. 

It is worth noting that the performance difference between the DiffusionVL here and DiffusionVL in~\cref{sec: exp_p1} does not imply that finetuning from AR-VLMs to dVLMs is the best approach. The DiffusionVL in~\cref{sec: exp_p1} benefits more from the fact that its base model has already undergone extensive, high-quality vision-language alignment training. We believe that AR-LMs, with longer and higher-quality visual finetuning, also have the potential to build dVLMs achieving performance comparable to dVLMs built from AR-VLMs on downstream benchmarks.

\subsection{Inference Speed and Quality}

\begin{figure*}[t]
    \centering
    \begin{subfigure}[t]{0.48\linewidth}
        \centering
        \includegraphics[width=\linewidth]{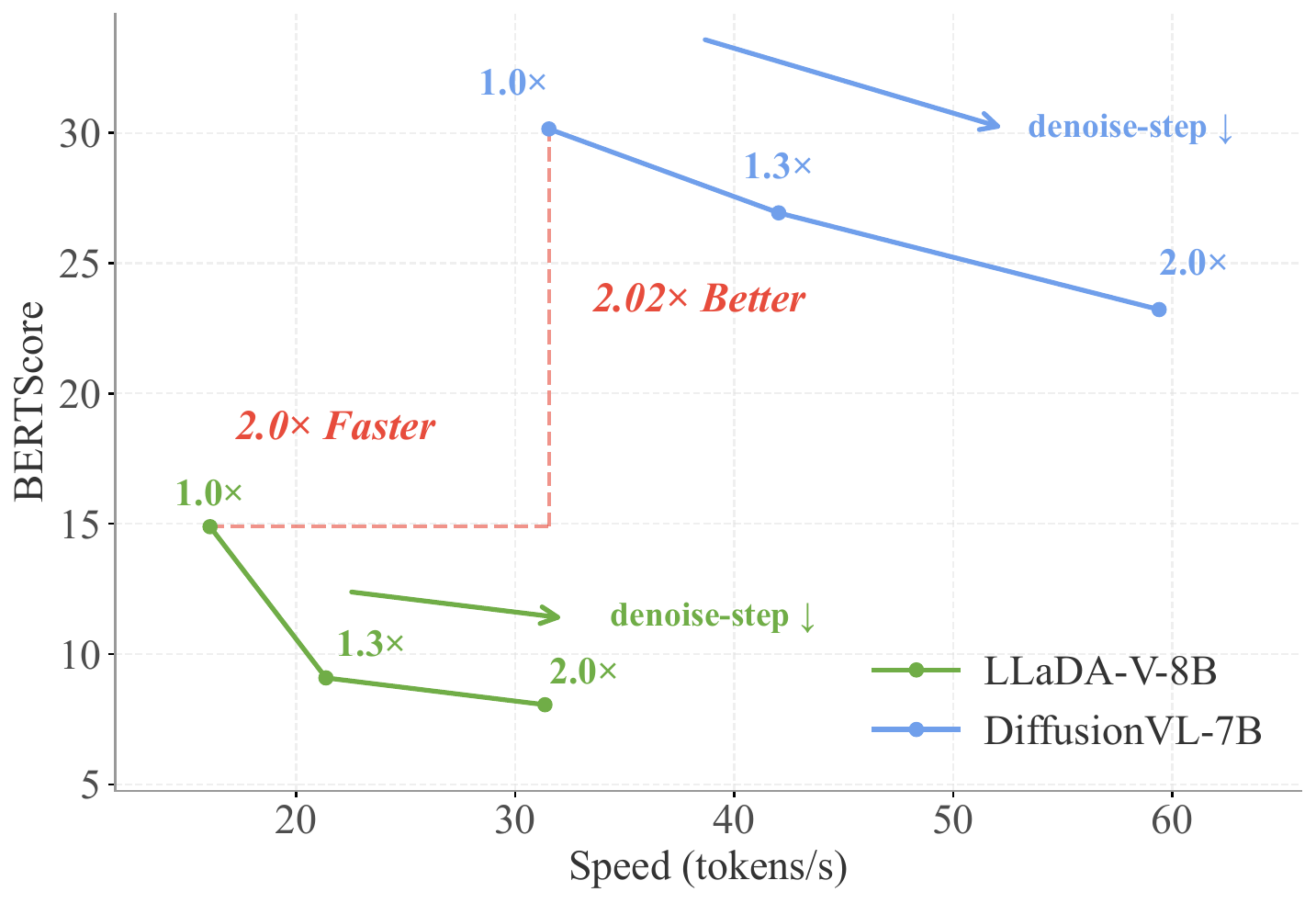}
        \caption{DetailCaps captioning performance.}
        \label{fig:detailcaps}
    \end{subfigure}
    \hfill
    \begin{subfigure}[t]{0.48\linewidth}
        \centering
        \includegraphics[width=\linewidth]{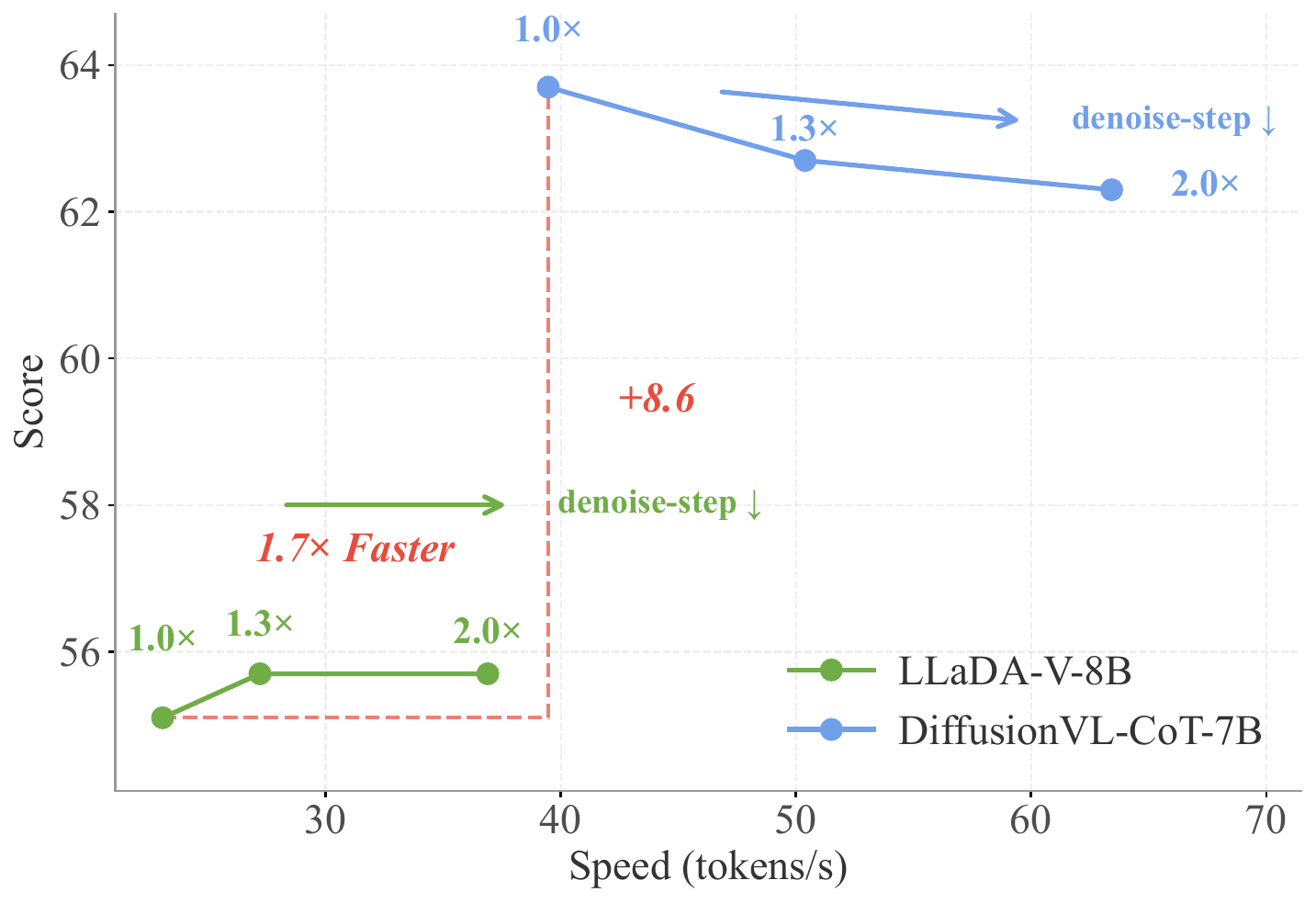}
        \caption{MathVista benchmark performance.}
        \label{fig:mathvista}
    \end{subfigure}
    \caption{\textbf{Speed and quality trade-offs on detailed captioning and math reasoning.} We define the parallelism factor for dVLMs as the average number of tokens generated simultaneously throughout the sequence (e.g., $1\times$ corresponds to single-token sampling). Speed metrics were collected on 8 GPUs and reported as the average per device.}
    \label{fig:combined_speed_quality}
\end{figure*}

\begin{table}[t]
    \centering
    \small
    \setlength{\tabcolsep}{2.8pt}
    \renewcommand{\arraystretch}{1.1}
    \begin{tabular*}{\columnwidth}{@{\extracolsep{\fill}}lcccccc@{}}
    \toprule
    \multirow{2}{*}{\textbf{Model}} &
    \multicolumn{2}{c}{\textbf{MathVista}} &
    \multicolumn{2}{c}{\textbf{MathVerse}} &
    \multicolumn{2}{c}{\textbf{MathVision}} \\
    \cmidrule(lr){2-3} \cmidrule(lr){4-5} \cmidrule(lr){6-7}
    & Acc.$\uparrow$ & TPS$\uparrow$ & Acc.$\uparrow$ & TPS$\uparrow$ & Acc.$\uparrow$ & TPS$\uparrow$ \\
    \midrule
    \thename{}-7B & \textbf{64.30} & 5.21 & \textbf{36.29} & 14.38 & \textbf{19.08} & 17.74 \\
    LLaDA-V-8B & 55.10 & \underline{23.07} & 31.35 & \underline{23.18} & 17.11 & \underline{22.75} \\
    \thename{}-CoT-7B & \underline{63.70} & \textbf{39.47} & \underline{32.74} & \textbf{42.31} & \underline{18.75} & \textbf{39.26} \\
    \bottomrule
    \end{tabular*}
    \caption{\textbf{Comprehensive comparison on math reasoning benchmarks.} We report both accuracy (Acc.) and inference throughput (TPS). Best results are in \textbf{bold} and second-best are \underline{underlined}.}
    \label{tab:comprehensive_comparison}
\end{table}

To evaluate inference speed and its trade-off with quality, we evaluate DiffusionVL on two representative tasks: detailed image captioning on DetailCaps \cite{dong2024benchmarking} and mathematical reasoning on multiple benchmarks, including MathVista \cite{lu2023mathvista}, MathVision \cite{ahmad2024mathvision}, and MathVerse \cite{zhang2024mathverse}. During inference, we employ the static low-confidence remasking strategy to control the number of decoded tokens at each denoising step. For both tasks, we compare against LLaDA-V-8B \cite{you2025llada} under its recommended fast-dLLM caching \cite{wu2025fast} with recomputation every 32 steps.~\cref{fig:combined_speed_quality} presents the speed-quality trade-offs on DetailCaps and MathVista, while~\cref{tab:comprehensive_comparison} summarizes the full results across all math benchmarks.

We first examine image captioning on DetailCaps, limiting responses to 512 tokens and evaluating against ground-truth captions using BERTScore \cite{bert-score}.  As shown in~\cref{fig:combined_speed_quality}(a), our DiffusionVL-7B achieves a BERTScore $2.02\times$ higher than LLaDA-V-8B while enjoying $2.0\times$ faster inference under the same parallelism. Moreover, we observe a clear test-time compute scaling law: increasing the number of denoising steps improves descriptive performance at the cost of inference speed, highlighting a tunable trade-off for DiffusionVL.

Turning to mathematical reasoning, we encounter a different challenge. While our DiffusionVL achieves competitive accuracy, its TPS lags behind LLaDA-V. This limitation, also observed in LaViDA \cite{li2025lavida}, stems from the absence of chain-of-thought (CoT) data in our training corpus. To address this, we randomly sample 100K CoT examples from OpenMMReasoner-SFT \cite{zhang2025openmmreasoner} and finetune the model with the original dataset, yielding DiffusionVL-CoT. For LLaDA-V, we enable fast-dLLM caching with a generation length of 64 (longer lengths degrade accuracy), while DiffusionVL-CoT uses a maximum length of 1024 to accommodate longer reasoning chains. As~\cref{fig:combined_speed_quality}(b) illustrates, DiffusionVL-CoT-7B outperforms LLaDA-V-8B by $8.6$ points on MathVista while achieving $1.7\times$ faster inference. ~\cref{tab:comprehensive_comparison} further confirms these gains across all three math benchmarks.

\subsection{Ablation Study}
\label{sec: ablate}

\noindent\textbf{Different block size performance.} To further explore the impact of different block sizes on the performance of diffusion finetuning, we conduct four groups of diffusion finetuning ablation experiments based on the Qwen2.5VL-3B-Instruct, setting the training block size from 1 to 16. During inference, we set the number of denoising steps equal to the block size to ensure models with different block sizes use the same number of denoising steps when generating the same tokens.

As shown in~\cref{tab:ablation1_block}, smaller block sizes yield marginally better accuracy, while larger ones offer greater parallel potential and less computational overhead, leading to substantially higher throughput ($1.55\times$ from block size 1 to 16). Block size 8 achieves the best DetailCaps BERTScore and offers a good speed--quality balance, so we adopt it as our default configuration.

\noindent\textbf{Different noise performance.} To validate the effectiveness of our block-level noise scheduling strategy, we compare it against the conventional sequence-level noise used in prior dVLMs. In sequence-level noise, masking is applied uniformly across the entire sequence. In contrast, our block-level noise applies a consistent noise ratio to each block independently. This design better aligns with the block decoding process during inference. As shown in~\cref{tab:ablation2}, block-level noise generally outperforms sequence-level noise across multiple benchmarks.

\noindent\textbf{Different diffusion paradigms.} To determine the optimal diffusion paradigm for VLMs, we perform block diffusion finetuning and full diffusion finetuning from the same Qwen2.5VL-7B-Instruct. \cref{tab:ablation2} demonstrates that block diffusion yields superior performance on multimodal benchmarks. Furthermore,~\cref{tab:comprehensive_comparison} shows that \thename{}-CoT (block diffusion) consistently achieves higher throughput than LLaDA-V (full diffusion), highlighting the inference efficiency advantage of the block diffusion paradigm. These results indicate that block diffusion strikes a superior balance between quality and efficiency, making it the preferred paradigm for building dVLMs from AR models.

\noindent\textbf{Different thresholds for dynamic remasking.} To explore more extreme acceleration, we adopt the dynamic low-confidence remasking strategy and conduct an ablation study on the relationship between the BERTScore / tokens per second (TPS) on DetailCaps and different thresholds under the condition of a fixed denoising step of 8. The choice of setting the denoising step to 8 is intended to minimize the number of tokens denoised in each fixed-schedule denoising step, which allows for a more intuitive demonstration of the differences between the dynamic remasking strategy and the static remasking strategy.

\cref{tab:ablation1_remask} shows that smaller dynamic thresholds allow the model to decode all tokens meeting the threshold condition at each denoising step, thereby achieving more significant acceleration. However, such acceleration comes at the cost of a certain degree of performance degradation for the model.

\begin{table}[t]
    \centering
    \setlength{\tabcolsep}{5pt}
    \renewcommand{\arraystretch}{1.0}
    \begin{tabular}{lcccc}
    \toprule
    \multicolumn{5}{c}{\textit{Training block size}} \\
    \midrule
    \textbf{Benchmark} & $b\!=\!1$ & $b\!=\!4$ & $b\!=\!8^\dagger$ & $b\!=\!16$ \\
    \midrule
    ChartQA         & \textbf{82.8} & 80.9 & 79.9 & 77.4 \\
    MMMU-Pro (std.) & \textbf{32.1} & 31.4 & 31.2 & 31.0 \\
    MMMU-Pro (vis.) & \textbf{22.2} & 19.9 & 20.2 & 18.1 \\
    MMStar          & \textbf{57.8} & 56.5 & 55.9 & 55.2 \\
    DetailCaps      & 29.8 & 29.9 & \textbf{30.9} & 29.9 \\
    \midrule
    \rowcolor{gray!14} TPS (tok/s) & 26.9 & 38.8 & 39.0 & \textbf{41.6} \\
    \bottomrule
    \end{tabular}
    \caption{\textbf{Ablation on training block size.} Effect of training block size on benchmark scores and throughput (TPS). The TPS throughput row is shaded in light gray. $^\dagger$\, denotes our default block size. Best per row in \textbf{bold}. TPS measured per GPU on DetailCaps.}
    \label{tab:ablation1_block}
\end{table}

\begin{table}[t]
    \centering
    \setlength{\tabcolsep}{4pt}
    \renewcommand{\arraystretch}{1.05}
    \begin{tabular}{lccc}
    \toprule
    \textbf{Metric} & \textbf{Sequence} & \textbf{Full} & \textbf{Block$^\dagger$} \\
    \midrule
    MMMU-Pro (std.) & 36.30 & 36.47 & \textbf{36.94} \\
    MMMU-Pro (vis.) & 24.86 & 19.88 & \textbf{24.97} \\
    MMStar          & 61.86 & 62.08 & \textbf{63.18} \\
    ChartQA         & 83.88 & 52.00 & \textbf{84.20} \\
    MME (cog.)      & 670.4 & 660.7 & \textbf{675.4} \\
    MME (perp.)     & \textbf{1527} & \textbf{1607} & 1519 \\
    \bottomrule
    \end{tabular}
    \caption{\textbf{Ablation on noise strategy and diffusion paradigm} (Qwen2.5VL-7B-Instruct). \textit{Block$^\dagger$} is the default block-wise setting; \textit{Sequence} and \textit{Full} are sequence-level noise and full diffusion, respectively. $^\dagger$\, denotes our default. Best in each pairwise comparison against Block is in \textbf{bold}.}
    \label{tab:ablation2}
\end{table}

\begin{table}[t]
    \centering
    \setlength{\tabcolsep}{8pt}
    \renewcommand{\arraystretch}{1.0}
    \begin{tabular}{lcccc>{\columncolor{gray!14}}c}
    \toprule
    \multicolumn{6}{c}{\textit{Dynamic remasking}} \\
    \midrule
    \textbf{$\tau$} & 1.0$^\dagger$ & 0.8 & 0.6 & 0.4 & 0.2 \\
    \midrule
    \textbf{TPS}$\uparrow$ & 31.5 & 34.6 & 37.8 & 46.6 & \textbf{71.8} \\
    \textbf{BERT}$\uparrow$ & \textbf{30.2} & 30.1 & 29.7 & 27.4 & 18.6 \\
    \bottomrule
    \end{tabular}
    \caption{\textbf{Ablation on dynamic remasking.} Dynamic low-confidence remasking at varying thresholds with fixed 8 denoising steps. $^\dagger$\, denotes our default. Best per group in bold. TPS measured per GPU on DetailCaps.}
    \label{tab:ablation1_remask}
\end{table}

\section{Conclusion and Future Work}

\noindent\textbf{Conclusion.} This paper focuses on a novel problem in the building of dVLMs: Is it possible to construct dVLMs based on existing powerful autoregressive models? In response, we propose \thename, a dVLM family that translates from any powerful autoregressive models. We obtain two key observations: (1) The paradigm shift from AR-VLMs to dVLMs is remarkably effective. (2) Direct conversion of an AR-LM to a dVLM is also feasible. Furthermore, we introduce a block decoding strategy into dVLMs that supports arbitrary-length generation and KV-cache reuse. With this integrated design, despite training with less than 5\% of the data required by prior methods, \thename~translated from AR-VLMs achieves state-of-the-art performance among existing dVLMs, alongside a $2.0\times$ inference speedup. \thename~translated from AR-LMs not only outperforms the dVLMs built from dLLMs but also achieves performance competitive with AR-VLMs finetuned under the same autoregressive paradigm.

\noindent\textbf{Future Work.} During our research on \thename{}, we observe that the AR-to-diffusion paradigm conversion is remarkably straightforward. We plan to explore whether this conversion paradigm can extend beyond finetuning to other training stages, including pretraining and reinforcement learning. Moreover, our inference experiments reveal that the diffusion paradigm yields significant efficiency gains in long chain-of-thought (CoT) scenarios, which motivates future work on the throughput benefits that \thename{} could bring to reinforcement learning.

{
    \small
    \bibliographystyle{ieeenat_fullname}
    \bibliography{main}
}

\end{document}